\documentclass{article}

\usepackage{graphicx}
\usepackage{bm}
\usepackage{authblk}
\usepackage[subtle]{savetrees}
\usepackage{biblatex}
\usepackage{amssymb}
\usepackage{amsmath}

\begin{document}
\date{}

\title{Reducing bias and increasing utility by federated generative modeling of medical images using a centralized adversary}

\author[1]{Jean-Francois Rajotte}
\author[2]{Sumit Mukherjee}
\author[2]{Caleb Robinson}
\author[2]{Anthony Ortiz}
\author[1]{Christopher West}
\author[2]{Juan M. Lavista Ferres}
\author[1]{Raymond T Ng}
\affil[1]{Data Science Institute, University of British Columbia}
\affil[2]{AI for Good Research Lab, Microsoft}
\maketitle


\begin{abstract}
A major roadblock in machine learning for healthcare is the inability of data to be shared broadly, due to privacy concerns. Privacy preserving synthetic data generation is increasingly being seen as a solution to this problem. However, since healthcare data often has significant site-specific biases, it has motivated the use of federated learning when the goal is to utilize data from multiple sites for machine learning model training.
Here, we introduce FELICIA (FEderated LearnIng with a CentralIzed Adversary), a generative mechanism enabling collaborative learning.
It is a generalized extension of the (local) PrivGAN mechanism allowing to take into account the diversity (non-IID) nature of the federated sites.
In particular, we show how a site with limited and biased data could benefit from other sites while keeping data from all the sources private.
FELICIA works for a large family of Generative Adversarial Networks (GAN) architectures including vanilla and conditional GANs as demonstrated in this work.
We show that by using the FELICIA mechanism, a site with a limited amount of images can generate high-quality synthetic images with improved utility, while none of the sites need to provide access to their real data.
The sharing happens solely through a central discriminator with access limited to synthetic data.
We demonstrate these benefits on several realistic healthcare scenarios using benchmark image datasets (MNIST, CIFAR-10) as well as on medical images for the task of skin lesion classification.
We show that the utility of synthetic images generated by FELICIA surpasses that of the data available locally and we demonstrate that it can correct the reduced utility of a biased subgroup within a class.
\end{abstract}


\section{Introduction}
\label{sec:intro}
Learning from images to build diagnostic or prognostic models of a medical condition has become a very active research topic because of its great potential to provide better care for patients.
Deep learning has been involved in much of modern progress in medical computer vision techniques, such as disease detection and classification as well as biomedical segmentation \cite{Esteva2021}. However, for such methods
to capture the subtle patterns between a medical condition and an image, it is important that a model is exposed to a rich variety of cases.
It is well known that images from a single source can be significantly biased by the demographics, equipment, and acquisition protocol~\cite{wachinger2021detect}.
Consequently, training a model on images from a single source would skew the performance of its prediction power towards the population from that source and potentially perform poorly for other populations.
Ideally, such a model should be trained on images from as many sources as possible.
To reduce the associated cost of collecting and labelling data, it is obvious that all sites such as hospitals and research centers would benefit to share their images.

Gaining access to large medical datasets requires a very lengthy approval process due to concerns about privacy breaches.
Most current privacy legislation prevents datasets from being accessed and analyzed outside of a small number of dedicated servers (e.g., servers within a local hospital). However, to unleash the full power of various machine learning techniques, particularly deep learning methods, we need to find ways to share data among research groups, while satisfying privacy requirements.

How sharing actually happens can depend on many factors such as use cases, regulation, business value protection and infrastructure availability.
In this work, we focus on synthetic data creation which allows multiple downstream use cases and exploration.
Our objective is to show how different \textit{sites} (e.g. hospitals) can help each other by creating joint or disjoint synthetic datasets that contain more utility than any of the single datasets alone. Moreover, the synthetic dataset can be used as a benchmark for machine learning in health care. To this end, we first test our method in two toy setups using common benchmark datasets, where we create artificial \textit{sites} with datasets from different data distributions. This shows the potential of our method in the domain of medical imaging.

\begin{figure}[ht]

\includegraphics[width=0.95\linewidth]{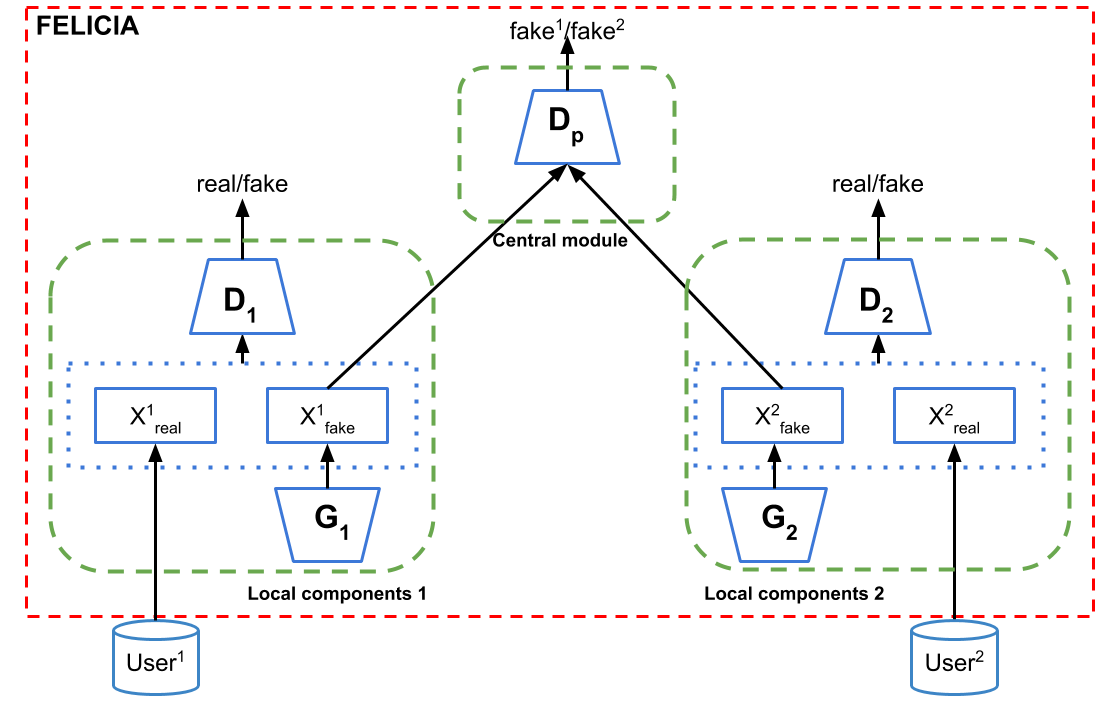}
\caption{FELICIA architecture with N=2 users. The real data subsets are determined by the users to which we associate local components by subscripts.
In this work, the users will often be referred to as \textit{sites} or hospitals in our experiment scenarios.}
\label{fig:arch_overview}

\label{fig:arch}
\end{figure}

Sharing private data or their characteristics has been extensively explored recently. A common approach is to generate privacy preserving synthetic data using various variants of Generative Adversarial Networks ~(GAN \cite{goodfellow2014generative}). GANs are generative models that are able to create realistic-looking synthetic images.
A GAN comprises of a generator \textbf{G} and a discriminator \textbf{D} playing a two-player game.
The generator aims to create fake samples such that the discriminator will estimate their probability to be as high as possible.
The discriminator on the other hand, tries to estimate the probability that a sample is real (rather than fake).
PrivGAN (\cite{mukherjee2021privgan}) is an extension of GAN, originally designed to generate synthetic data while improving the privacy of the data used for training.
Although PrivGAN was developed to be applied locally on a single dataset, previous work~(\cite{rajotte2020private}) has demonstrated that PrivGAN can be useful in a federated learning setting.
In this paper, we develop a general mechanism (FELICIA) to extend a large family of GANs to a federated learning setting utilizing a centralized adversary.
We explore the application of this framework to show how different sites can collaborate with each other to improve machine learning models in a privacy-preserving distributed data sharing scenario.
To demonstrate the relevance of FELICIA, we focus on settings relevant to health care. However, other natural scenarios can be found in disparate domains such as in banking.

Our main contributions are the following:
\begin{itemize}
    \item Formalize a new federated learning mechanism (FELICIA) motivated by the PrivGAN (\cite{mukherjee2021privgan}) architecture, which extends to a family of GAN architectures.
    \item Demonstrate empirically that the hyperparameter \(\lambda\) can improve the utility, contrary to the original PrivGAN.
    \item Generalize the hyperparameter \(\lambda\) to be site-dependent.
    \item Improve the synthetic data by using generators from multiple epochs as was done in ~(\cite{Beaulieu-Jones159756}).
    \item Demonstrate the applicability of FELICIA on real non-IID data both for conditional and non-conditional synthetic data generation.
    \item Demonstrate the applicability of using FELICIA to enable medical images sharing in a federated learning context.
    \item Demonstrate that FELICIA can create a synthetic dataset without the utility bias from its local data. 
\end{itemize}

\section{Related work}

Sharing data between non-local sites such as hospitals and research centers can be achieved in many ways.
A popular approach to share data with privacy is to generate private synthetic data with Differential Privacy ~(\cite{10.1561/0400000042}).
Generative models such as GANs based on either differentially private stochastic gradient descent ~(\cite{Abadi_2016,xie2018differentially}) or the Private Aggregation of Teacher Ensembles, PATE~(\cite{papernot2016semisupervised,yoon2018pategan}) are of particular interest.
Both approaches suffer from low utility data for a reasonable degree of privacy.

Another approach is to train a model in a federated learning setting such that the data never has to be shared (\cite{Rieke2020,rasouli2020fedgan,hardy2018mdgan,fan2020federated}).
Since it has been demonstrated that GANs are vulnerable to privacy attacks ~(\cite{hayes2017logan}), various approaches have been proposed to provide better privacy protection.
Synthetic data from GANs trained on distributed datasets with differential privacy ~(\cite{geyer2017differentially,chen2020gswgan}) suffer from the same low quality as synthetic data from centrally trained GANs, unless they have access to a very large amount of training data as in this language model application (\cite{mcmahan2017learning}).
FELICIA allows users to create high quality local synthetic datasets while privacy protection naturally arises from the architecture.


\subsection{PrivGAN}
PrivGAN (\cite{mukherjee2021privgan}) is an extension of a GAN originally designed to protect against membership inference attacks, such as LOGAN and MACE (\cite{hayes2017logan,liu2020mace}).
The architecture is comprised of $N$ GANs trained on disjoint, independent and identically distributed (IID) subsets with an extra loss from a central (private) discriminator \textbf{D\textsubscript{P}}.
The authors show that their method ``\emph{minimally affects the quality of downstream samples as evidenced by the performance on downstream learning tasks such as classification}''.
The key feature here is that the only connection between the subsets is the central discriminator \textbf{D\textsubscript{P}} accessing only synthetic data.

\section{Methods}

\subsection{Mathematical formulation}

While the original formulation of privGAN can be seen as a modification to the original GAN architecture (\cite{goodfellow2014generative}),
the mechanism of utilizing multiple generator-discriminator pairs and a centralized adversary is quite general. To that end, we first define a general family of GANs (\cite{arora2017generalization}) that contain a single generator \(G\),
a single discriminator \(D\) and a loss governed by a measure function
\(\phi : [0,1] \longrightarrow \mathbb{R}\) as follows:

\begin{align}\label{eqn:gen}
    \begin{split}
        V_{\phi}(G,D) =& \mathbb{E}_{x\sim p_{data}(x)}[\phi(D(x))]+ \\
    &\mathbb{E}_{z\sim p_{z}(z)}[\phi(1-D(G(z)))]
    \end{split}
\end{align}

In the case of conditional GANs, \(x\) is replaced by the conditioned tuple \((x|y)\) where \(y\) is the label associated with sample \(x\). 
Our proposed mechanism (FELICIA) extends any GAN belonging to this family to a federated learning setting using a centralized adversary. Formally, given a measure function $\phi$ and corresponding GAN loss $V_{\phi}$, the federated loss is: 
\begin{align}
\label{eqn:felicialoss}
    \text{FELICIA}(\phi,V_{\phi},\bm{\lambda},N) = \sum_{i=1}^N \underbrace{V_{\phi} (G_i,D_i)}_{local} + \lambda_{i} \underbrace{ R^{i,\phi}_p(D_p)}_{global}
\end{align}
Where \(R^{i,\phi}_p(D_p) = \mathbb{E}_{z\sim p_{z}(z)}\phi[D_p^i(G_i(z))]\).
A notable novelty here is that \(\bm{\lambda}\) is now a \(N\)-dimensional parameter \(\bm{\lambda} = (\lambda_1,...,\lambda_N)\), one for each of the \(N\) user.
Contrary to PrivGAN, both terms in FELICIA's loss have the potential to contribute to utility : \textit{local} favors utility on local data and \textit{global} favors utility on all users' data.

In this paper, we apply our mechanism to three separate GANs belonging to this family: i) the original GAN~(\cite{goodfellow2014generative}), ii) DCGAN~(\cite{radford2015unsupervised}), and iii) conditional GAN~(\cite{mirza2014conditional}).
We note however, that these are simply representative examples and the mechanism applies to a wide variety of GANs such as WGAN~(\cite{arjovsky2017wasserstein}), DP-GAN~(\cite{xie2018differentially}), etc.

\subsection{Practical implementation}
To implement the FELICIA mechanism we follow a process similar to the original PrivGAN paper.
Specifically, we duplicate the discriminator and generator architectures of a `base GAN' to each of the component generator-discriminator pairs of FELICIA.
The privacy discriminator (\textbf{D\textsubscript{P}}) is selected to be identical in architecture to the other discriminators barring the activation of the final layer.
Most of the optimization effort is dedicated to train the base GAN on the whole training data to generate realistic images.
Then FELICIA's implementation is optimized with the base GAN's parameters which are tuned to get good looking samples.
This last step is usually much faster as the base GAN's parameters represent a good starting point.

\section{Experiments}
Our experiments are based on a simulation of two hospitals (\textbf{Hospital 1} and \textbf{Hospital 2}) with different patient populations.
We consider a regulation preventing sharing images as well as models that had access to images.
We will use FELICIA where \textbf{Hospital 1} and \textbf{Hospital 2} correspond respectively to \textit{User\textsuperscript{1}} and \textit{User\textsuperscript{2}} in Figure \ref{fig:arch_overview}.
For our last two experiments, we define the concept of \textit{helpee} and \textit{helper}.
The \textit{helpee} is a hospital with low utility and biased dataset and the \textit{helper} is a hospital with a rich and high utility dataset willing to help within the above regulation restrictions.
We will show that, through the FELICIA framework, the \textit{helpee}, \textbf{Hospital 1}, can locally generate a less unbiased synthetic dataset with more utility than its own (real) data.

First, we use the MNIST dataset~(\cite{lecun2010mnist}) to show how FELICIA can help generate synthetic data with better coverage of the input distribution, \textit{even when both sites have a biased coverage of the possible input space}.
Second, we use a more complex dataset, CIFAR-10, to show how the utility could be significantly improved when a subgroup is underrepresented in the data.
Finally, we test FELICIA in a federated learning setting with medical imagery using a skin lesion image dataset.
In the first experiment, the utility is demonstrated visually by showing the distribution of the generated samples.
In the other experiments, the utility is defined as the performance of a classifier trained on synthetic data (sometimes combined with real data) and evaluated on a held out real dataset.

In the first two experiments, we have kept the default parameters of the original PrivGAN implementation, namely equal \(\lambda\)'s (i.e. \(\lambda_1 = \lambda_2 = 1\)).
We have also used the generator at the end of the training phase which gave satisfactory results.
In our last experiment however, such implementation did not lead to synthetic data with satisfactory utility.
We hypothesized that user-dependent \(\lambda_i\)'s would better suit the scenario of a \textit{helpee} being more penalized when its synthetic data is distinguishable from the \textit{helper}'s synthetic data.
Conversely, the \textit{helper} which can generate good quality synthetic data on its own, does not need to be penalized as much when it's synthetic data is distinguishable from a the biased \textit{helpee}.
Also, we have noticed that the utility does not increase asymptotically with increasing epoch.
Inspired by previous work~(\cite{Beaulieu-Jones159756}), we used synthetic images from 5 generators from the top epochs in utility.
The selection of the top epochs (as well as the best combination of (\(\lambda_1, \lambda_2\)), was determined with the hold out validation set and the final utility was determined on the test set.

\subsection{Improving distribution coverage}
One setting that multiple sites may observe is when \textbf{Hospital 1} owns a dataset with samples from one part of the input distribution, while \textbf{Hospital 2} has a dataset with samples from a different part.
We simulate this setting using the 28x28 gray scale hand written digit dataset MNIST~(\cite{lecun2010mnist}). Specifically, we test whether FELICIA is able to generate representative samples from the entire input distribution while the local data is biased.

Given all images of a selected digit, we perform PCA and cluster the images in the resulting embedding space using K-means (\(k=2\)). The resulting clusters will be used to distribute the images to the \textit{sites}.

We then train FELICIA using a varying proportion of images from both clusters and compare the resulting generated images to the original images.
We also compare with images generated by traditional GANs trained only on data from cluster 1 and cluster 2.
Specifically, we define a mixing parameter, $\alpha$, used to select the number of samples from each cluster used to fit FELICIA and two simple GANs.
FELICIA will be trained on the two subsets defined as follows:
\begin{description}
    \item[Subset 1] A random selection of $\alpha\%$ of samples from cluster 1, and a fraction $(100-\alpha)\%$ of samples from cluster 2.
    \item[Subset 2] Same as Subset 1 but inverting the fraction, i.e. replacing $\alpha\%$ by $(100-\alpha)\%$.
\end{description}
\textbf{Subset 1} and \textbf{Subset 2} correspond respectively to \textbf{X\textsuperscript{1}\textsubscript{Real}} and \textbf{X\textsuperscript{2}\textsubscript{Real}} in the right diagram of Figure \ref{fig:arch_overview}.

For $\alpha=0$, \textbf{Subset 1} will be completely biased towards cluster 2 (representing a specific section of the input distribution), and for $\alpha=50$ both subsets will consist of equal numbers of samples spread over the input distribution.
FELICIA's training results in a generator for \textbf{Subset 1}, \textbf{G\textsubscript{1}}, and generator for \textbf{Subset 2}, \textbf{G\textsubscript{2}}.
We also train two simple GANs using the data from \textbf{Subset 1} and \textbf{Subset 2} respectively.

\begin{figure}[!ht]
\includegraphics[width=0.95\linewidth]{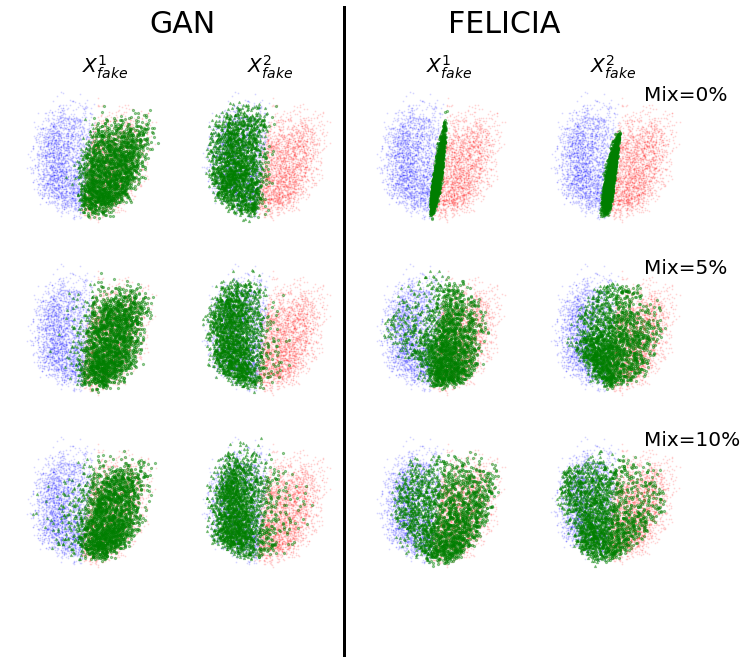}
\centering
\caption{GAN and FELICIA generated samples on biased subsets for digit four.
Each points corresponds to the position of a given hand written digit in the first two component of the PCA embedding.
As a reference, each plot shows clusters 1 (blue points) and cluster 2 (red points) in the background.
The green points correspond to the generated samples.
Column 1 \& 2 show simple GAN generated samples after training on subset 1 \& 2, respectively.
Column 3 \& 4 show generated samples by FELICIA \textbf{G\textsubscript{1}} and \textbf{G\textsubscript{2}}, respectively.}
\label{fig:mnist_cluster_mix_embedded_digit4}
\end{figure}

Once all GANs are trained, we generate 2000 samples from each and compare them to the original samples by plotting all samples using the first two principal components from the original image embedding step.
Figure \ref{fig:mnist_cluster_mix_embedded_digit4} shows such plots for three values of $\alpha$.
When the bias is maximal (i.e. when $\alpha = 0$), FELICIA generates images only at the cluster border, while the simple GANs will generate images only from the cluster on which they were trained.
This is not surprising when we consider that if the local discriminator \textbf{D\textsubscript{1}} is never trained on real images from a given cluster -- it will not ``allow'' the generator to cover that part of the input space -- the only generated samples that satisfy both discriminators are those at the border.
As $\alpha$ increases (shown in descending rows of Figure \ref{fig:mnist_cluster_mix_embedded_digit4}), it is clear that the samples generated by FELICIA cover more of the input space than those of the local GANs.

\subsection{Reducing bias}
\label{subsec:cifar}

Another setting that various sites may observe is when one owns an imbalanced dataset while the other owns a complete (unbiased) dataset.
In this setting, the owner of the imbalanced dataset, the \textit{helpee}, should be able to benefit from the owner with a balanced dataset, the \textit{helper}.
We use the CIFAR-10 dataset~\cite{krizhevsky2009learning} to simulate such a setting.
CIFAR-10 is a dataset of 32x32 RGB images labeled with 10 different classes of animals and transport vehicles.
To represent a biased dataset, we define two classes: \textsc{class 1}, the house pets class, consisting of ``cats'' and ``dogs'' and \textsc{class 2}, the large animal class, consisting of ``deers'' and ``horses''.
Similarly to the previous experiment, we will create two subsets:
\begin{description}
    \item[Subset 1] Contains an equal number of cat and dog samples for \textsc{class 1} and an unequal number of deer and horse samples for \textsc{class 2}. The bias of this subset will be quantified with $\beta$, the fraction of \textsc{class 2} samples that are images of deer. This represents the \textit{helpee}'s dataset.
    \item[Subset 2] Contains an equal number of cat and dog samples for \textsc{class 1} and an equal number of deer and horse samples for \textsc{class 2}. This represents the \textit{helper}'s dataset.
\end{description}
Note that the two subsets have an equal number of images; the difference is in the proportion of deers \& horses of the samples that make up \textsc{class 2}.

We train a CNN to discriminate between \textsc{class 1} and \textsc{class 2} with three different training sets: \textbf{Subset 1} only, \textbf{Subset 1} + GAN synthetic data (i.e. augmented with GAN), and \textbf{Subset 1} + FELICIA synthetic (i.e. augmented with FELICIA) data, then measure the classification accuracy on a held out test set.
FELICIA synthetic data is created from the \textit{helpee}'s generator associated to \textbf{Subset 1}.

\begin{figure}[!ht]
\centering
\includegraphics[width=0.45\linewidth]{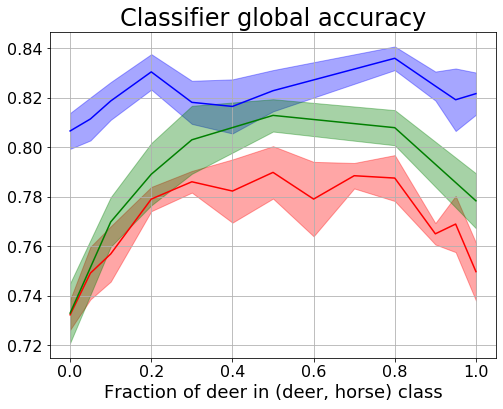}
\includegraphics[width=0.45\linewidth]{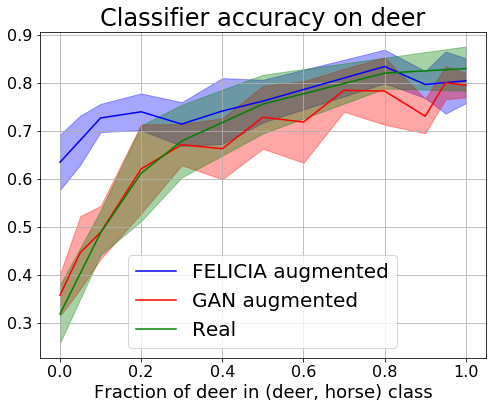}
\caption{Classifier accuracy on three data subsets after training on real data, real data augmented with synthetic data from a simple GAN and real data augmented with FELICIA. (\textbf{Left}) accuracy over samples from all classes. (\textbf{Right}) accuracy over samples of the deer class.
}
\label{fig:cifar_results}
\end{figure}

Figure \ref{fig:cifar_results} shows the accuracy as function of $\beta$ of each classification model evaluated on the full held out test set and its deer images only.
We observe that the classifier accuracy after training on real data decreases when the data is more biased towards the deer.
This is expected as the test data is balanced and the reduced subgroup in the training set leads to reduced accuracy.
This is confirmed in the right panels of Figure \ref{fig:cifar_results}, showing a decrease in accuracy of the classifier consistent with the biais of the training data.
The same figure shows that augmenting the classifier training set with simple GANs synthetic data does not improve the accuracy.
This is also expected as a simple GAN goal is to reproduce the training data distribution.
Finally, the classifier trained on real data augmented with FELICIA synthetic data is systematically better than other classifiers.
The improvement is particularly significant when the data is most biased.

\subsection{Synthetic medical images}
In our last experiment, we apply FELICIA in a federated learning setting with a real-world medical image dataset, HAM10000~(\cite{tschandl2018ham10000}).
Similar to the previous experiment, we will use these images to simulate a biased subset for the \textit{helpee} \textbf{Hospital 1}.
This dataset contains a large collection of multi-source dermatoscopic images of common pigmented skin lesions\footnote{The original images have a size of 600x450 pixels that we have resized to 64x64 pixels in order to train models more quickly.}.
These are separated into 7 imbalanced sets of skin lesion images, from which we use the four most populated, \textsc{lesion 0}: Melanocytic nevi (6705 images), \textsc{lesion 1}: Melanoma (1113 images), \textsc{lesion 2}: Benign keratosis (1099 images) \textsc{lesion 3}: Basal cell carcinoma (514 images).
Lesions 0 and 2 are benign whereas lesions 1 and 3 are associated with different types of skin cancer.
We create two classes from these lesion sets: 
\begin{description}
    \item[Class 0] Images of benign lesions from \textsc{lesion 0} \& \textsc{lesion 2}
    \item[Class 1] Images for cancerous lesions from \textsc{lesion 1} \& \textsc{lesion 3}
\end{description}

We evaluate the performance of a binary classifier trained to predict whether a lesion is benign (\textsc{class 0}) or cancerous (\textsc{class 1}).
This type of skin lesion classification has shown to be successful with deep learning (see ~(\cite{Esteva2021}) and references therein).

We first randomly remove 1000 images from each class to create two equal size held out sets: a test and a validation set. 
From the remaining dataset, the first subset is defined similarly as the previous experiment with balanced classes but artificially biased in the lesion within one of the class.
The training subsets are defined as follows:
\begin{description}
    \item[Subset 1] Contains 300 images from each classes.
    \begin{itemize}
        \item \textsc{Class 0} (benign) biased with 10 images of \textsc{lesion 0} and 290 images \textsc{lesion 2}.
        \item \textsc{Class 1} (cancerous) balanced with 150 images of \textsc{lesions 1} and 150 images from \textsc{lesion 3}.
    \end{itemize}
    \item[Subset 2] Contains the remainder of the dataset.
\end{description}
Where \textbf{Subset 1} and \textbf{Subset 2} correspond respectively to \textbf{X\textsuperscript{1}\textsubscript{Real}} and \textbf{X\textsuperscript{2}\textsubscript{Real}} in the FELICIA diagram in Figure \ref{fig:arch_overview}.

We explore how the \textit{helpee} (\textbf{Subset 1}) with limited and biased data, could be helped by the \textit{helper}'s (\textbf{Subset 2}) richer data through the FELICIA mechanism.
In this experiment, the utility is defined as the performance of a benign/cancerous classifier trained on synthetic data and evaluated on the held out set.
Specifically, we use area under the receiver operator characteristic curve (AUC-ROC).
Since we are dealing with two hyperparameters (\(\lambda_1, \lambda_2\)), they are selected with the validation set and the final performance is reported on the test set.
To address the relatively low amount of images (compared to the previous experiments), we use a conditional GAN~(\cite{mirza2014conditional}) to leverage all training images for one model as oppose to a model per class as in the previous experiments\footnote{For simplicity, we have kept the central discriminator \(D_p\) \textit{unconditional}, i.e. without class input.}.

We train FELICIA on the two subsets over 100000 epochs for various combinations of (\(\lambda_1, \lambda_2\)) from equation (\ref{eqn:felicialoss}).
For each set of (\(\lambda_1, \lambda_2\)), we re-run the experiment multiple times while varying the random seeds for the network initialization and data shuffling for the train, test and validation set.

For comparison, we train a conditional GAN using data \textit{only} from \textbf{Subset 1}.
This represents the synthetic data that the \textit{helpee} could generate without access to the \textit{helper}'s dataset.

We evaluate the utility of the generated images every 50 training epochs.
For each evaluated epoch, we used the generator to create 200 images for each class.
A simple CNN classifier trained on these generated images is then evaluated on the 500 images of the balanced validation set.
Then we select the best combination of (\(\lambda_1, \lambda_2\)) and make our final evaluation on a held out test set.

\begin{figure}[ht]
\centering
\includegraphics[width=0.95\linewidth]{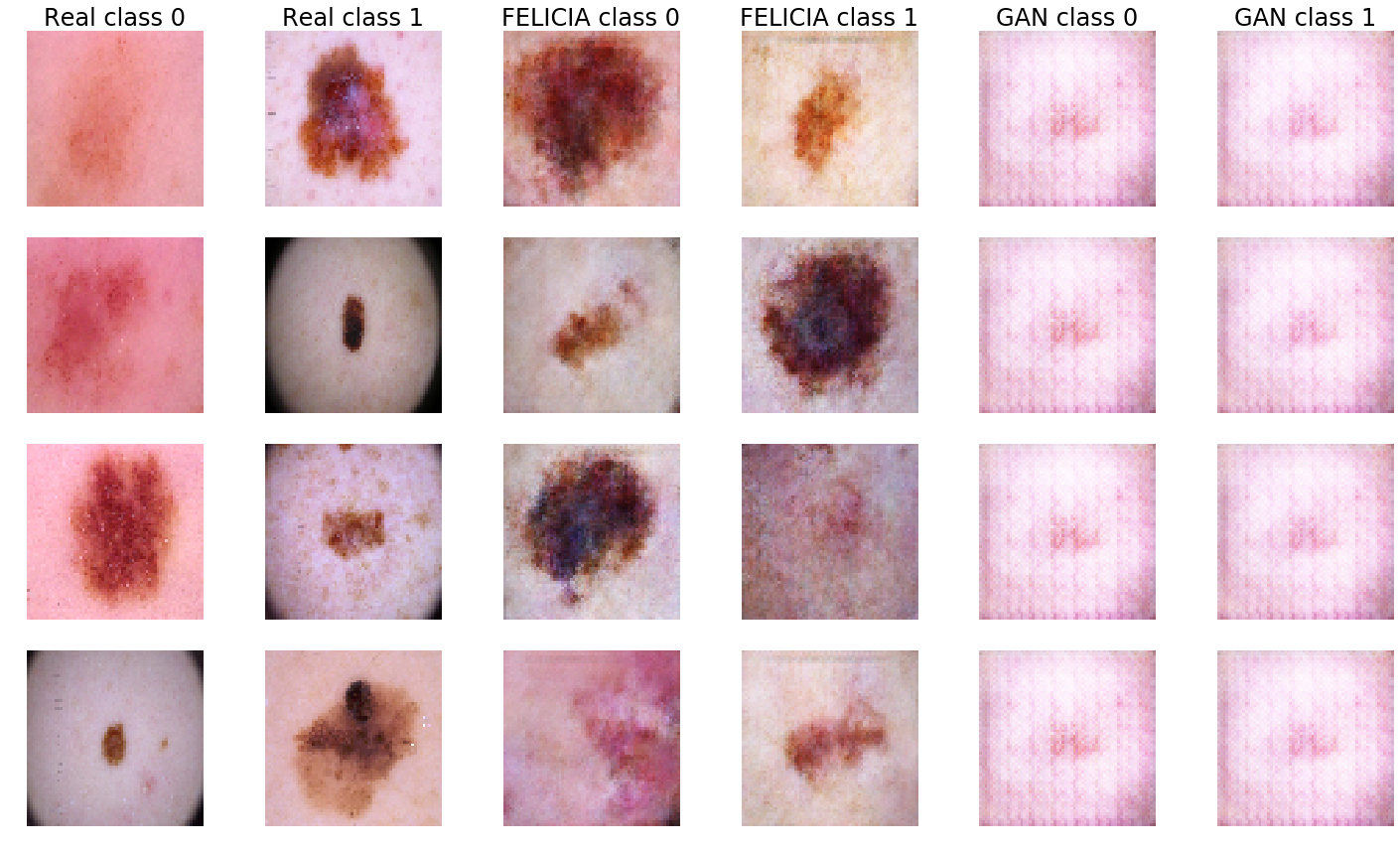}
\caption{Real images and synthetic images generated by FELICIA and conditional GAN. The images are created by the generator from the epoch of highest utility.}
\label{fig:skin_sample}
\end{figure}

\begin{figure}[ht]
\centering
\includegraphics[width=0.95\linewidth]{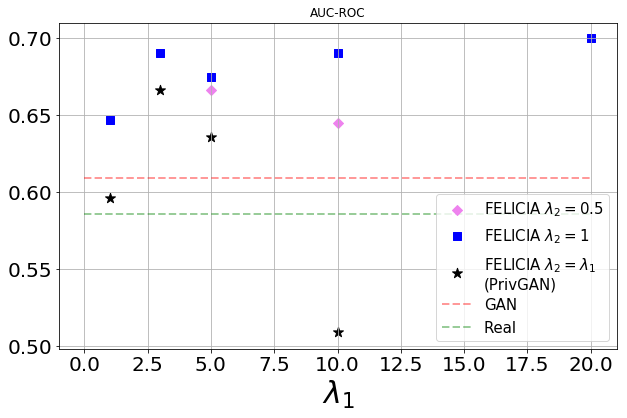}
\caption{\textit{Helpee} synthetic images utility as function of the \(\lambda_1\) parameter, i.e. the strength of the loss to the central discriminator applied to the \textit{Helpee} generator total loss.}
\label{fig:skin_score_vs_lambda1}
\end{figure}

Figure \ref{fig:skin_sample} shows how the \textit{helpee} can generate very realistic images with FELICIA while images from the simple GAN are of very low quality and lack diversity.
The images are produced from the saved generator at epoch leading to the best utility.
We show in Figure \ref{fig:skin_score_vs_lambda1} the utility as function of several combination of \((\lambda_1, \lambda_2)\).
For simple (conditional) PrivGAN, we limit the search to \(\lambda_1 = \lambda_2\) and a single generator as in the original paper~(\cite{mukherjee2021privgan}).
We see that PrivGAN improves less the overall utility and for a limited range of \(\lambda\). 

Table \ref{tab:skin_scores} summarized the utility metrics for the synthetic data generated by the \textit{helpee}. FELICIA data surpasses in utility both the local real data (\textbf{Subset 1}) and the synthetic data from a simple (conditional) GAN.
Furthermore, Table \ref{tab:skin_scores} shows how FELICIA improves the accuracy of the utility classifier for the penalized subgroup (\textit{Melanocytic nevi}) without significantly affecting the other subgroups.
\\
\begin{table}[ht]

\begin{tabular}{lcccc}

Metric                          & REAL  & GAN   & PrivGAN & FELICIA \\
    & & & \begin{tiny} FELICIA \(\lambda_1 = \lambda_2\)\end{tiny}\\
\hline
AUC-ROC                         & 0.59 & 0.61 & 0.67    & 0.69    \\
Accuracy (Melanocytic nevi)     & 0.51 & 0.74 &     0.71 &     0.73    \\
Accuracy (Benign keratosis)     & 0.57 & 0.65 &     0.65 &     0.65    \\
Accuracy (Melanoma)             & 0.55 & 0.39 &     0.47 &     0.51    \\
Accuracy (Basal c. carcinoma)   &  0.74 & 0.52 &     0.62 &     0.65    \\
\end{tabular}%
\caption{Utility of skin lesion synthetic images generated by the \textit{helpee}.
\textit{Melanocytic nevi} is the biased subgroup (equivalent to the \textit{deer} in the previous section).
We note that FELICIA both improves the overall score while keeping the performance on subgroups more balanced.}
\label{tab:skin_scores}
\end{table}

\subsection{Discussion}

Our experiments suggest that FELICIA allows generators to learn distributions beyond a local subset. This is supported by the clusters coverage in the first experiment on the hand written digit as well as the improved utility of a synthetic dataset compared to the local data.
Moreover, these results could not be reproduced with synthetic data produced by a GAN on the same local data.

We note that in our last experiment FELICIA is selecting the \(\lambda_i\) hyperparameters based on the evaluation on a holdout dataset.
In an application, this data could be available as a public dataset.
When this is not possible, the utility could be determined by training a classifier on synthetic data and be evaluated securely on the \textit{helper}'s data with differential privacy, e.g. ~(\cite{10.1145/3313276.3316377}, \cite{10.1145/2808769.2808775}).
Alternatively, one might be interested in using the \(\lambda\)'s to weight the client's contribution for diversity rather than overall performance.
In that case, they could be equal unless some sites hold very small datasets, their  \(\lambda\)'s would then have to be reduced (presumably) proportionally to the datasets size.

\section{Conclusions}
\label{sec:conclusions}
We have developed a novel mechanism, FELICIA, that allows for the sharing of data more securely in order to generate synthetic data in a federated learning context.
By setting up various scenarios with biased sites, we have demonstrated the advantages of our mechanism with image datasets.
We have shown that a biased site can be securely helped by another site through the FELICIA architecture and that it will benefit more the more biased it is.
We have also demonstrated on medical images that FELICIA can help generate synthetic images with more utility than images available locally.
The work presented was implemented centrally, therefore the performance effect of the sites being distributed is still to be investigated.

FELICIA can be implemented with a wide variety of GANs which will depend on the type of data and use case.
A particularly relevant use case is a pandemic such as COVID-19 where hospitals and research centers at the beginning of an outbreak would benefit from the data gathered by sites affected earlier.
The data sharing approval process can easily take months, whereas pandemic microbiology evolution tells us that a virus can mutate to a different strain orders of magnitude faster.
Another application is the augmentation of an image dataset to improve diagnostic such as the classification of cancer pathology images (\cite{Levine2020.02.24.963553}).
The data from one research center is often biased towards some dominating population of the available data for training.
FELICIA could help mitigate such bias by allowing sites from all over the world to create a synthetic dataset based on a more general population.

We are currently working on implementing FELICIA with progressive GAN in order to generate highly complex medical images such as CT scans, x-rays and histopathology slides in a real federated learning setting with non-local sites.

\section*{Acknowledgement}
We are grateful to the Cascadia Data Discovery Initiative for enabling this collaboration and for granting Azure credits for part of this work.

\printbibliography

@article{mukherjee2021privgan,
  title={privGAN: Protecting GANs from membership inference attacks at low cost to utility},
  author={Mukherjee, Sumit and Xu, Yixi and Trivedi, Anusua and Patowary, Nabajyoti and Ferres, Juan L},
  journal={Proceedings on Privacy Enhancing Technologies},
  volume={3},
  pages={142--163},
  year={2021}
}

@inproceedings{hardy2018mdgan,
  title={{MD-GAN}: Multi-discriminator generative adversarial networks for distributed datasets},
  author={Hardy, Corentin and Le Merrer, Erwan and Sericola, Bruno},
  booktitle={2019 IEEE International Parallel and Distributed Processing Symposium (IPDPS)},
  pages={866--877},
  year={2019},
  organization={IEEE}
}

@article{10.1561/0400000042,
  title={The algorithmic foundations of differential privacy.},
  author={Dwork, Cynthia and Roth, Aaron and others},
  journal={Foundations and Trends in Theoretical Computer Science},
  volume={9},
  number={3-4},
  pages={211--407},
  year={2014}
}

@article{rasouli2020fedgan,
  title={{FedGAN}: Federated generative adversarial networks for distributed data},
  author={Rasouli, Mohammad and Sun, Tao and Rajagopal, Ram},
  journal={arXiv preprint arXiv:2006.07228},
  year={2020}
}

@inproceedings{Abadi_2016,
  title={Deep learning with differential privacy},
  author={Abadi, Martin and Chu, Andy and Goodfellow, Ian and McMahan, H Brendan and Mironov, Ilya and Talwar, Kunal and Zhang, Li},
  booktitle={Proceedings of the 2016 ACM SIGSAC conference on computer and communications security},
  pages={308--318},
  year={2016}
}

@article{papernot2016semisupervised,
  title={Semi-supervised knowledge transfer for deep learning from private training data},
  author={Papernot, Nicolas and Abadi, Mart{\'\i}n and Erlingsson, Ulfar and Goodfellow, Ian and Talwar, Kunal},
  journal={arXiv preprint arXiv:1610.05755},
  year={2016}
}

@inproceedings{yoon2018pategan,
    title={{PATE}-{GAN}: Generating Synthetic Data with Differential Privacy Guarantees},
    author={Jinsung Yoon and James Jordon and Mihaela van der Schaar},
    booktitle={International Conference on Learning Representations},
    year={2019}
}

@article{xie2018differentially,
  title={Differentially private generative adversarial network},
  author={Xie, Liyang and Lin, Kaixiang and Wang, Shu and Wang, Fei and Zhou, Jiayu},
  journal={arXiv preprint arXiv:1802.06739},
  year={2018}
}

@article{geyer2017differentially,
  title={Differentially private federated learning: A client level perspective},
  author={Geyer, Robin C and Klein, Tassilo and Nabi, Moin},
  journal={arXiv preprint arXiv:1712.07557},
  year={2017}
}

@article{mcmahan2017learning,
  title={Learning differentially private recurrent language models},
  author={McMahan, H Brendan and Ramage, Daniel and Talwar, Kunal and Zhang, Li},
  journal={arXiv preprint arXiv:1710.06963},
  year={2017}
}

@inproceedings{fan2020federated,
  title={Federated generative adversarial learning},
  author={Fan, Chenyou and Liu, Ping},
  booktitle={Chinese Conference on Pattern Recognition and Computer Vision (PRCV)},
  pages={3--15},
  year={2020},
  organization={Springer}
}

@article{chen2020gswgan,
  title={{GS-WGAN}: A gradient-sanitized approach for learning differentially private generators},
  author={Chen, Dingfan and Orekondy, Tribhuvanesh and Fritz, Mario},
  journal={arXiv preprint arXiv:2006.08265},
  year={2020}
}

@misc{lecun2010mnist,
  title={MNIST handwritten digit database},
  author={LeCun, Yann and Cortes, Corinna and Burges, CJ},
  howpublished={\url{http://yann.lecun.com/exdb/mnist}},
  year={2010}
}

@article{hayes2017logan,
  title={{LOGAN}: Membership inference attacks against generative models},
  author={Hayes, Jamie and Melis, Luca and Danezis, George and De Cristofaro, Emiliano},
  journal={arXiv preprint arXiv:1705.07663},
  year={2017}
}

@article{liu2020mace,
  title={MACE: A Flexible Framework for Membership Privacy Estimation in Generative Models},
  author={Liu, Xiyang and Xu, Yixi and Mukherjee, Sumit and Ferres, Juan Lavista},
  journal={arXiv preprint arXiv:2009.05683},
  year={2020}
}

@article{krizhevsky2009learning,
  title={Learning multiple layers of features from tiny images},
  author={Krizhevsky, Alex and Hinton, Geoffrey and others},
  year={2009},
  publisher={Citeseer}
}

@article{tschandl2018ham10000,
  title={The HAM10000 dataset, a large collection of multi-source dermatoscopic images of common pigmented skin lesions},
  author={Tschandl, Philipp and Rosendahl, Cliff and Kittler, Harald},
  journal={Scientific data},
  volume={5},
  number={1},
  pages={1--9},
  year={2018},
  publisher={Nature Publishing Group}
}

@article{goodfellow2014generative,
  title={Generative adversarial networks},
  author={Goodfellow, Ian J and Pouget-Abadie, Jean and Mirza, Mehdi and Xu, Bing and Warde-Farley, David and Ozair, Sherjil and Courville, Aaron and Bengio, Yoshua},
  journal={Advances in Neural Information Processing Systems},
  year={2014},
  publisher={Curran}
}

@inproceedings{arora2017generalization,
  title={Generalization and equilibrium in generative adversarial nets ({GAN}s)},
  author={Arora, Sanjeev and Ge, Rong and Liang, Yingyu and Ma, Tengyu and Zhang, Yi},
  booktitle={International Conference on Machine Learning},
  pages={224--232},
  year={2017},
  organization={PMLR}
}

@inproceedings{rajotte2020private,
  title={Private data sharing between decentralized users through the privGAN architecture},
  author={Rajotte, Jean-Francois and Ng, Raymond T},
  booktitle={2020 IEEE 24th International Enterprise Distributed Object Computing Workshop (EDOCW)},
  pages={37--42},
  year={2020},
  organization={IEEE}
}

@article{radford2015unsupervised,
  title={Unsupervised representation learning with deep convolutional generative adversarial networks},
  author={Radford, Alec and Metz, Luke and Chintala, Soumith},
  journal={arXiv preprint arXiv:1511.06434},
  year={2015}
}

@inproceedings{arjovsky2017wasserstein,
  title={Wasserstein generative adversarial networks},
  author={Arjovsky, Martin and Chintala, Soumith and Bottou, L{\'e}on},
  booktitle={International Conference on Machine Learning},
  pages={214--223},
  year={2017},
  organization={PMLR}
}

@article{Levine2020.02.24.963553,
  title={Synthesis of diagnostic quality cancer pathology images by generative adversarial networks},
  author={Levine, Adrian B and Peng, Jason and Farnell, David and Nursey, Mitchell and Wang, Yiping and Naso, Julia R and Ren, Hezhen and Farahani, Hossein and Chen, Colin and Chiu, Derek and others},
  journal={The Journal of pathology},
  volume={252},
  number={2},
  pages={178--188},
  year={2020},
  publisher={Wiley Online Library}
}

@article{Esteva2021,
  title={Deep learning-enabled medical computer vision},
  author={Esteva, Andre and Chou, Katherine and Yeung, Serena and Naik, Nikhil and Madani, Ali and Mottaghi, Ali and Liu, Yun and Topol, Eric and Dean, Jeff and Socher, Richard},
  journal={NPJ digital medicine},
  volume={4},
  number={1},
  pages={1--9},
  year={2021},
  publisher={Nature Publishing Group}
}

@article{Rieke2020,
  title={The future of digital health with federated learning},
  author={Rieke, Nicola and Hancox, Jonny and Li, Wenqi and Milletari, Fausto and Roth, Holger R and Albarqouni, Shadi and Bakas, Spyridon and Galtier, Mathieu N and Landman, Bennett A and Maier-Hein, Klaus and others},
  journal={NPJ digital medicine},
  volume={3},
  number={1},
  pages={1--7},
  year={2020},
  publisher={Nature Publishing Group}
}

@article{mirza2014conditional,
  title={Conditional generative adversarial nets},
  author={Mirza, Mehdi and Osindero, Simon},
  journal={arXiv preprint arXiv:1411.1784},
  year={2014}
}

@inproceedings{10.1145/3313276.3316377,
  title={Private selection from private candidates},
  author={Liu, Jingcheng and Talwar, Kunal},
  booktitle={Proceedings of the 51st Annual ACM SIGACT Symposium on Theory of Computing},
  pages={298--309},
  year={2019},
  organization={ACM}
}

@inproceedings{10.1145/2808769.2808775,
  title={Differential privacy for classifier evaluation},
  author={Boyd, Kendrick and Lantz, Eric and Page, David},
  booktitle={Proceedings of the 8th ACM Workshop on Artificial Intelligence and Security},
  pages={15--23},
  year={2015},
  organization={ACM}
}

@article{Beaulieu-Jones159756,
  title={Privacy-preserving generative deep neural networks support clinical data sharing},
  author={Beaulieu-Jones, Brett K and Wu, Zhiwei Steven and Williams, Chris and Lee, Ran and Bhavnani, Sanjeev P and Byrd, James Brian and Greene, Casey S},
  journal={Circulation: Cardiovascular Quality and Outcomes},
  volume={12},
  number={7},
  year={2019},
  publisher={Am Heart Assoc}
}

@article{wachinger2021detect,
  title={Detect and correct bias in multi-site neuroimaging datasets},
  author={Wachinger, Christian and Rieckmann, Anna and P{\"o}lsterl, Sebastian and Alzheimer’s Disease Neuroimaging Initiative and others},
  journal={Medical Image Analysis},
  volume={67},
  pages={101879},
  year={2021},
  publisher={Elsevier}
}
\newpage
\appendix
\section{Supplement material}

\begin{figure}[h]
\includegraphics[width=1.0\linewidth]{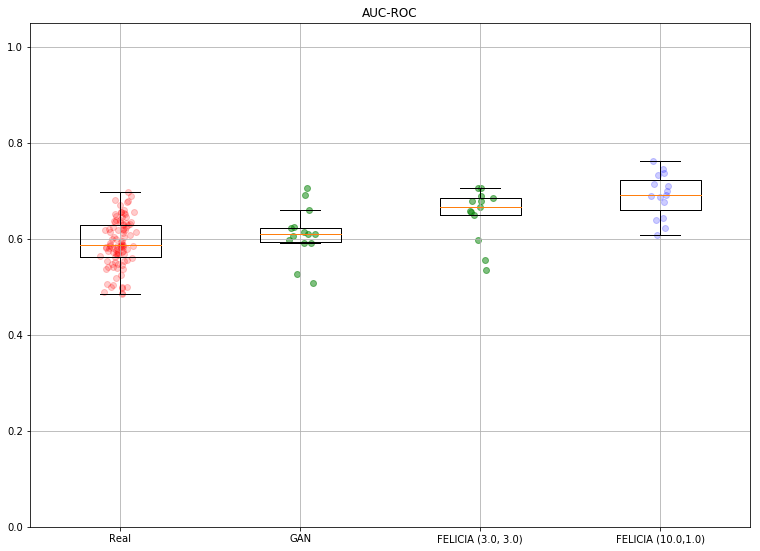}
\caption{Repeated AUC-ROC performance evaluation on the Real dataset (helpee with different seed for the classifier and shuffling) and the best synthetic lesion images from GAN, FELICIA (with PrivGAN parameters \(\lambda_1=\lambda_2\)) and FELICIA.
The values in bracket correspond to the selected (\(\lambda_1\), \(\lambda_2\)).
Each \textit{Real} point corresponds to training a classifier initialized with a different random seed on a training set selected with a different shuffling seed.
Each synthetic point (GAN or FELICIA) corresponds to a different random seed for the data generation and the shuffling for creating the (real) subsets.
The box extends from the lower to upper quartile values of the data, with a line at the median. The whiskers extend from the box to show the range of the data.}
\label{fig:auc_scatter}
\end{figure}

\begin{figure}[h]
\includegraphics[width=1.0\linewidth]{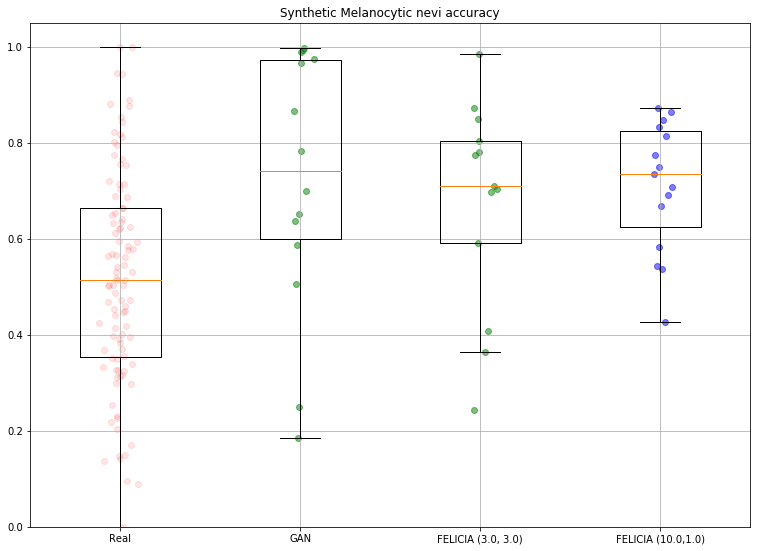}
\caption{Same as Figure \ref{fig:auc_scatter} but for accuracy on the Melanocytic nevi lesions.}
\label{fig:acc_meln}
\end{figure}

\begin{figure}[h]
\includegraphics[width=1.0\linewidth]{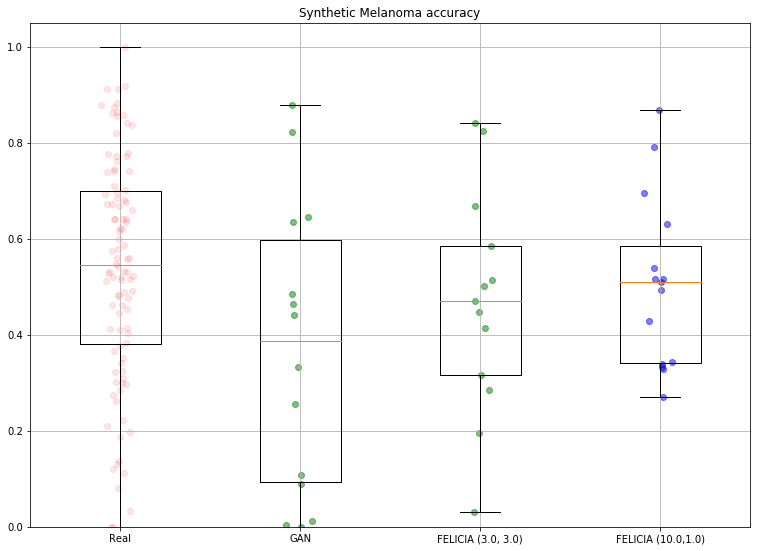}
\caption{Same as Figure \ref{fig:auc_scatter} but for accuracy on the Melanoma lesions.}
\label{fig:acc_mela}
\end{figure}

\begin{figure}[h]
\includegraphics[width=1.0\linewidth]{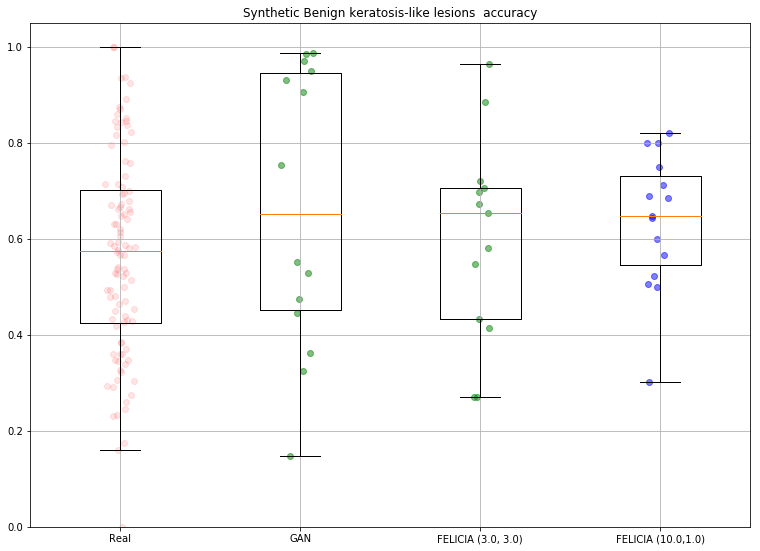}
\caption{Same as Figure \ref{fig:auc_scatter} but for accuracy on the Benign keratosis lesions.}
\label{fig:acc_kera}
\end{figure}

\begin{figure}[h]
\includegraphics[width=1.0\linewidth]{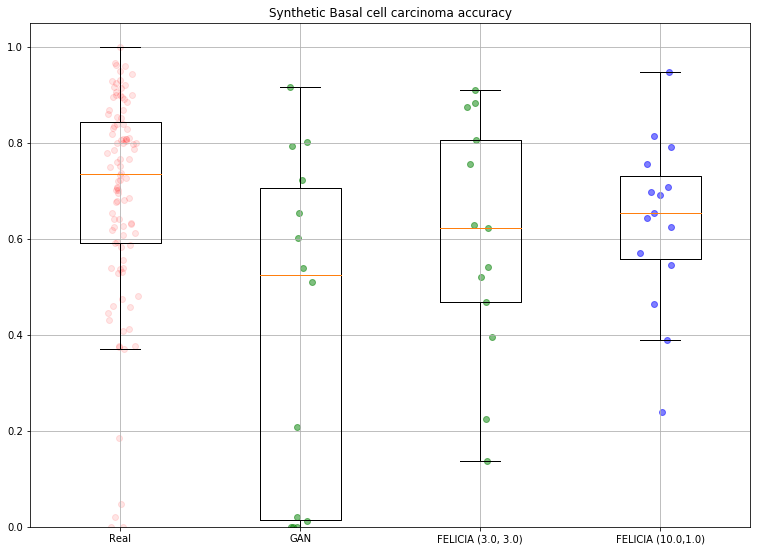}
\caption{Same as Figure \ref{fig:auc_scatter} but for accuracy on the Basal cell carcinoma lesions.}
\label{fig:acc_basa}
\end{figure}

We show here more details related to the Table \ref{tab:skin_scores}. Figures \ref{fig:auc_scatter}, \ref{fig:acc_meln}, \ref{fig:acc_kera}, \ref{fig:acc_mela}, \ref{fig:acc_basa} corresponds to the row 1-5 of Table 1. More precisely, the middle lines in the box plots (the medians) correspond to the values of Table 1. We note that, not only FELICIA has the best overall performance, but it also has the smallest performance scatter on the subgroups.

\end{document}